\def\year{2021}\relax
\documentclass[letterpaper]{article} 
\usepackage{aaai21}  
\usepackage{times}  
\usepackage{helvet} 
\usepackage{courier}  
\usepackage[hyphens]{url}  
\usepackage{graphicx} 
\usepackage{natbib}  
\usepackage{caption} 

\usepackage{subcaption}
\usepackage{algorithm,algpseudocode}
\usepackage{multirow}

\usepackage{amsmath,amssymb}

\usepackage{pifont}
\newcommand{\cmark}{\ding{52}}%
\newcommand{\xmark}{\ding{56}}%
\usepackage[switch]{lineno}

\title{Semantic Grouping Network for Video Captioning\footnote{\protect\url{https://github.com/hobincar/SGN}}}
\author{
    Hobin Ryu,
    Sunghun Kang,
    Haeyong Kang,
    and Chang D. Yoo
    \\
}
\affiliations{

    Korea Advanced Institute of Science and Technology (KAIST), Daejeon, Republic of Korea \\
    \{hobincar, sunghun.kang, haeyong.kang, cd\_yoo\}@kaist.ac.kr

}

\begin{document}

\maketitle

\begin{abstract}
This paper considers a video caption generating network referred to as Semantic Grouping Network (SGN) that attempts (1) to group video frames with discriminating word phrases of partially decoded caption and then (2) to decode those semantically aligned groups in predicting the next word.
As consecutive frames are not likely to provide unique information, prior methods have focused on discarding or merging repetitive information based only on the input video.
The SGN learns an algorithm to capture the most discriminating word phrases of the partially decoded caption and a mapping that associates each phrase to the relevant video frames - establishing this mapping allows semantically related frames to be clustered, which reduces redundancy.
In contrast to the prior methods, the continuous feedback from decoded words enables the SGN to dynamically update the video representation that adapts to the partially decoded caption. 
Furthermore, a contrastive attention loss is proposed to facilitate accurate alignment between a word phrase and video frames without manual annotations.
The SGN achieves state-of-the-art performances by outperforming runner-up methods by a margin of 2.1\%p and 2.4\%p in a CIDEr-D score on MSVD and MSR-VTT datasets, respectively. Extensive experiments demonstrate the effectiveness and interpretability of the SGN.
\end{abstract}

\section{Introduction}

Video captioning is the task of understanding the scenes in a video and describing it in words. It is one of the most challenging computer vision tasks as it requires a model capable of associating video to text.
Most video captioning methods have been suggested based on the encoder-decoder framework constructed using convolutional neural networks (CNNs) and recurrent neural networks (RNNs). The CNN-based encoder takes a set of consecutive frames of the input video and produces visual representations to generate the accurate caption that describes the video. Then, the RNN-based decoder takes the visually encoded features and the previously predicted word as input and generates the caption one word at a time.

Unlike image captioning, which requires a model to understand static content in a single image, video captioning requires a model to understand the comprehensive context of a video.
A video frame is similar to the previous frame, and consecutive frames usually do not provide unique information \cite{PickNet}.
Therefore, considering every frame as an independent unit of information is not an efficient way to understand the video.
It is quite natural for humans to understand a video by partitioning it into information units based on semantics.
We understand a video by grouping information based on meanings such as people, objects, or actions, rather than frame by frame.

\begin{figure}[t]
    \centering
    \includegraphics[width=0.5\textwidth]{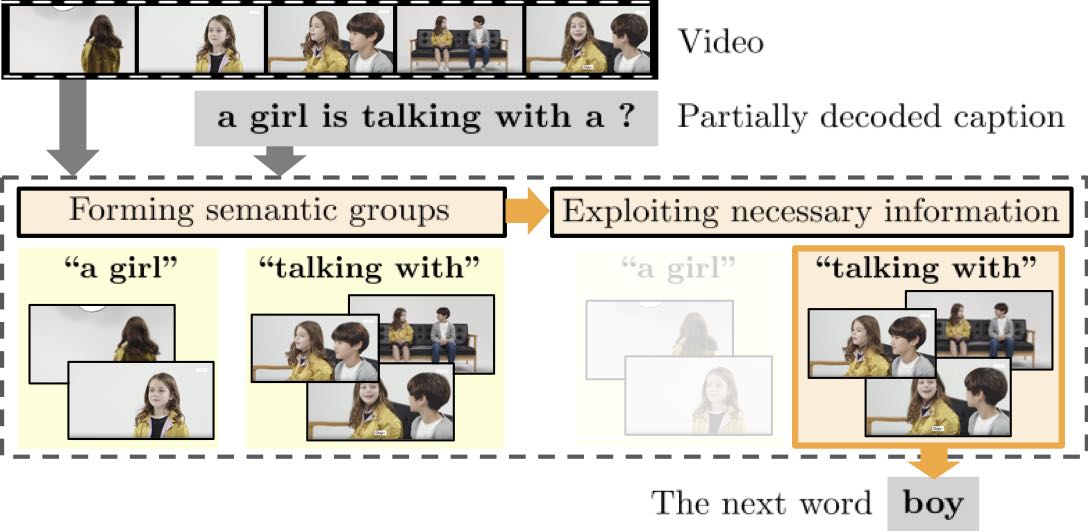}
    \caption{Phrases (e.g., ``a girl" and ``talking with") can gather their relevant frames in the video, forming groups that share common semantics within them. A decoder then exploits the necessary semantic group in predicting the next word (``boy").}
    \label{fig:intuition}
\end{figure}

There are a considerable number of works that try to imitate the human behavior of understanding video - collect the semantically related information into units and then decode the collected information units into a caption.
The type of information units varies; there are methods that partition a video into a fixed or adaptive number of segments that consist of successive frames \cite{HRNE,BAE}, collect the frames that are informative enough \cite{PickNet}, or gather all the features of video frames at the object-level \cite{OA-BTG,ORG-TRL,SAAT}.

\begin{figure*}[t]
    \centering
    \begin{subfigure}{.25\textwidth}
      \centering
      \includegraphics[width=.97\linewidth]{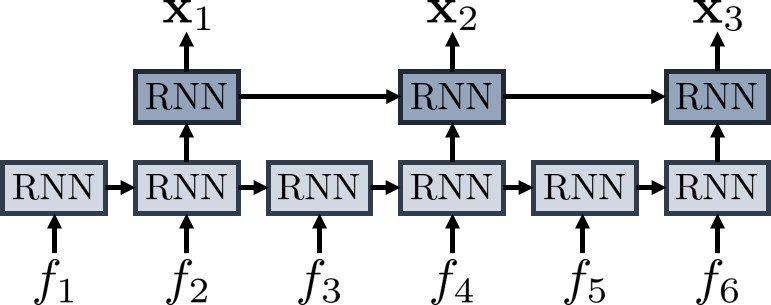}
      \caption{\small HRNE \cite{HRNE}}
      \label{fig:hrne}
    \end{subfigure}%
    \centering
    \begin{subfigure}{.25\textwidth}
      \centering
      \includegraphics[width=.97\linewidth]{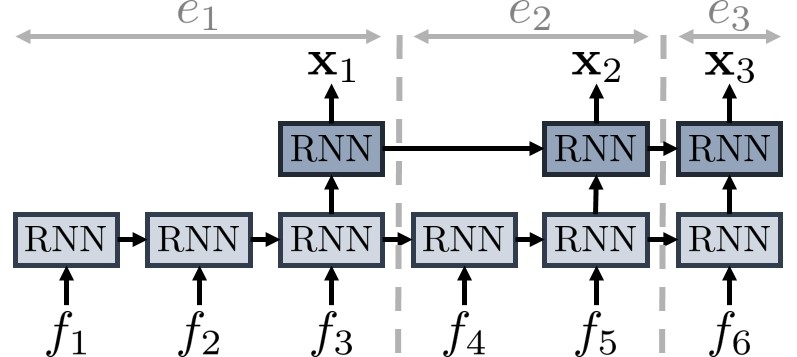}
      \caption{\small BAE (Baraldi et al. 2017)}
      \label{fig:bae}
    \end{subfigure}%
    \centering
    \begin{subfigure}{.25\textwidth}
      \centering
      \includegraphics[width=.97\linewidth]{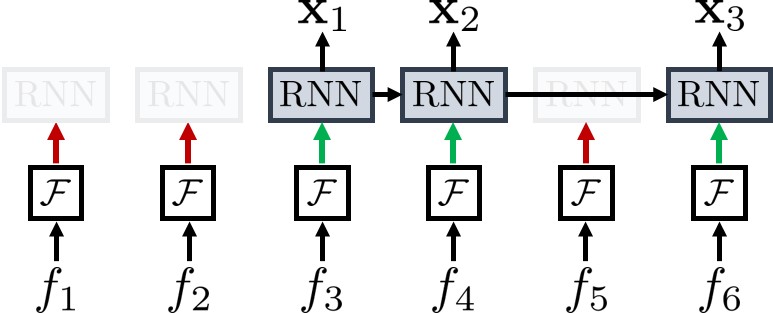}
      \caption{\small PickNet \cite{PickNet}}
      \label{fig:picknet}
    \end{subfigure}%
    \centering
    \begin{subfigure}{.25\textwidth}
      \centering
      \includegraphics[width=.97\linewidth]{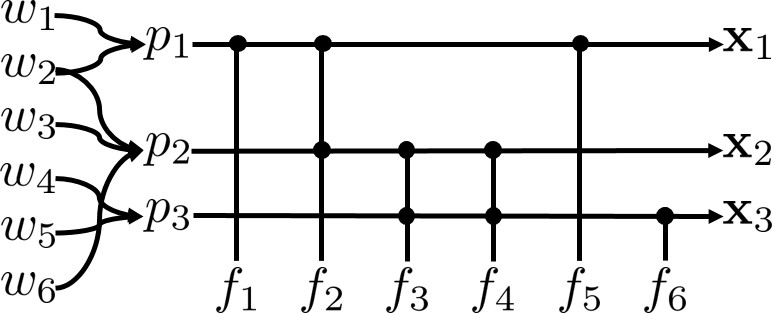}
      \caption{\small SGN (Ours)}
      \label{fig:SGN}
    \end{subfigure}
    \caption{Illustration showing the video encoding process: how video representations $\mathbf{x}$ are obtained in four different video captioning methods. $f$, $e$, $w$, $p$ denotes frames, events, words of the partially decoded caption, and phrases, respectively.}
    \label{fig:related}
\end{figure*}

It is important for a captioning method to model visual and textual modalities respectively and complementarily.
However, prior methods mainly focus on the visual aspect (i.e., video frames) and do not pay much attention to the textual aspect (i.e., partially decoded caption) when encoding a video.
The partially decoded caption, which consists of words predicted by the decoder, basically summarizes the visual content.
Therefore, the word phrases of the partially decoded caption could help associate semantically related frames into information units to form a group, which is referred to as \textit{semantic groups}.
For instance, let us consider a video shown in Figure \ref{fig:intuition} where a girl meets a boy and talks, and the decoder has partially generated ``a girl is talking with a".
The phrase ``a girl" can be used to group the first three frames where the girl is standing alone, and the last three frames can be grouped with the phrase ``talking with" as there are two people talking.
Given these two groups, the decoder can exploit the semantic meaning which these two groups represent in predicting the next word ``boy".

To use semantic groups as information units for understanding a video, they should satisfy the following three properties. 
First, the meaning of a semantic group should be concrete and observable.
When grouping frames based on semantics, a phrase is more suitable than a word which may have insufficient information to specify, such as function words like ``is" and ``the".
Second, a semantic group should have a meaning that is distinctive to others to effectively use it as a separate information unit without redundancy. 
Third, a phrase and corresponding frames in a semantic group should be semantically aligned to have a coherent meaning.
In other words, all the frames in a semantic group should be closely related to their phrase.

To this end, this paper proposes a video caption generating network referred to as \textbf{Semantic Grouping Network (SGN)}, which encodes a video into semantic groups by aligning frames around the phrases of partially decoded caption and describes the video by exploiting the semantic groups as information units.
SGN constructs phrases using words of partially decoded caption and forms a semantic group based on each phrase to which the frames are aligned.
By considering both input video features and partially decoded caption when further encoding the video, SGN can adaptively decode the next caption word depending on the caption already decoded. 
This capability is in contrast to prior methods which does not have any feedback control from partially decoded caption when encoding the video.
Also, the visual-textual alignment enables the captioning model to know the visual groundings of word phrases of the caption, which leads to a comprehensive understanding of the captioning context.
To further facilitate the correct semantic alignment within a semantic group, a \textbf{Contrastive Attention (CA) loss} is proposed to penalize the semantic groups that include some unrelated frames.

The key contributions of this work can be summarized as follows.
First, this paper proposes a Semantic Grouping Network (SGN), which encodes the video into semantic groups that are in terms of relevant frames and the corresponding word phrases of the partially decoded caption, and adaptively decodes the next word based on the semantic groups.
Second, this paper proposes Contrastive Attention (CA) loss that provides labor-free supervision for the correct visual-textual alignment within each semantic group.
Third, the SGN outperforms the runner-up methods by a large margin in CIDEr-D on the two most popular benchmark datasets, reaching new heights in state-of-the-art performances.

\section{Related Works}
\label{sec:related}

\textbf{Encoding Video into Information Units.}
\label{subsec:infounits}
As consecutive video frames contain highly repetitive information, several video captioning methods encode video into information units to imitate the human behavior of understanding video.
HRNE \cite{HRNE} employs a hierarchical LSTM to encode the input video into two-levels of abstraction. 
Later, BAE \cite{BAE}, which assumes video as a set of consecutive events, improves the HRNE by discovering the hierarchical structure of the video.
PickNet \cite{PickNet} assumes that all frames selected by equal interval sampling are not guaranteed to contain meaningful information and selects the only frames that are informative for describing the video.
These methods encode the input video by discarding or merging intermediate frames without considering the caption being generated; once the input video is encoded, the same video features are used throughout the decoding process. 
As SGN encodes the video based on the partially decoded caption by leveraging the word phrases to construct the \textit{semantic groups}, the video representation is adaptive to its own generated caption and the captioning model can better exploit the whole captioning context (see Figure \ref{fig:related}).

\begin{figure*}[t]
    \centering
    \includegraphics[width=0.96\textwidth]{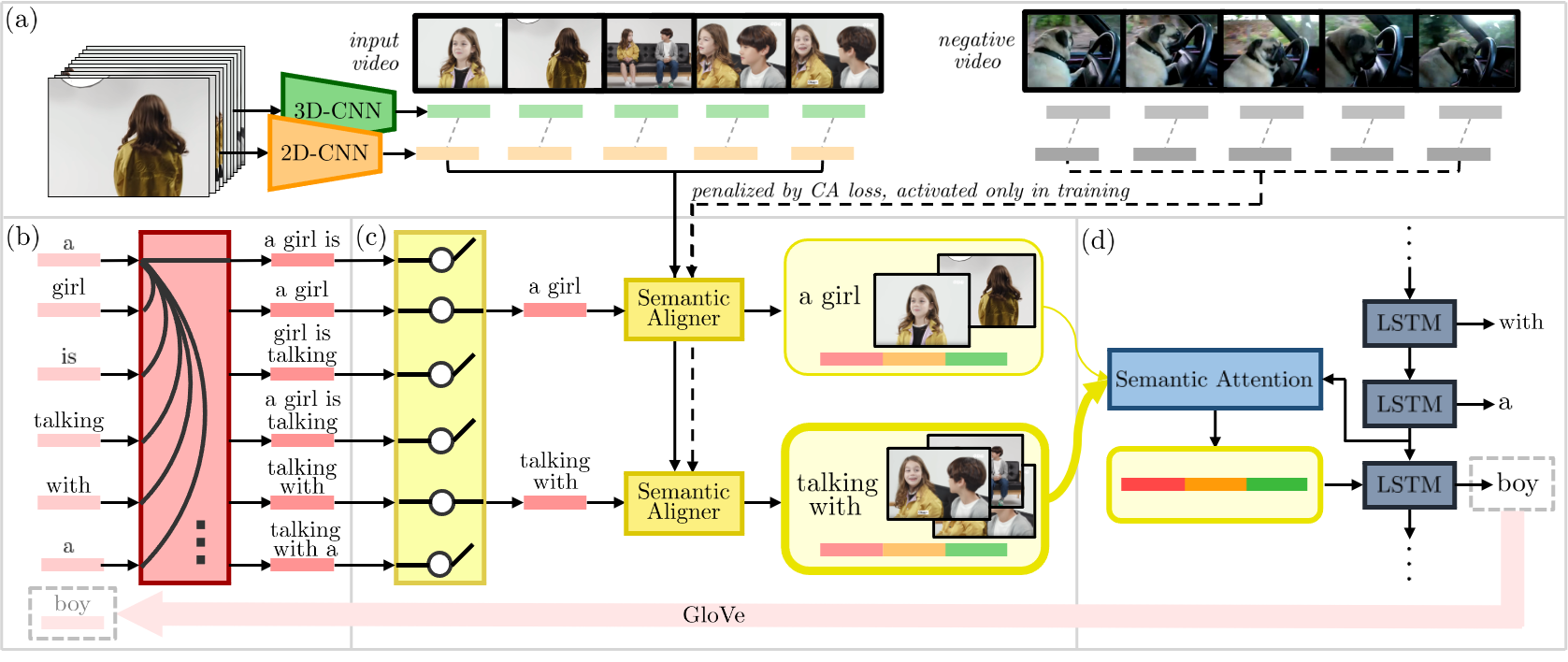}
    \caption{The SGN consists of (a) Visual Encoder, (b) Phrase Encoder, (c) Semantic Grouping, and (d) Decoder. In training, a negative video is introduced in addition to the input video for calculating the CA loss. The words predicted by the Decoder are added to the input of the Phrase Encoder and become word candidates that make up phrases.}
    \label{fig:architecture}
\end{figure*}

\textbf{Multi-Modal Reasoning.}
\label{subsec:multimodal}
As captioning involves both visual and textual modalities, understanding one with the help of the other is an interesting research area.
You et al. \cite{SemanticAttention} detect visual concepts in the image and leverage the word embedding of each concept as the key to its visual representation when applying an attention mechanism.
GLIED \cite{GLIED} argues that a more effective attention mechanism can be employed by considering the collocations of detected concepts along with their visual representations.
M3 \cite{M3} is equipped with a heterogeneous memory to model the long-term visual-textual dependency, and MARN \cite{MARN} is equipped with a memory consisting of words and corresponding visual contexts across videos to utilize the videos other than the input video.
Inspired by the potential of leveraging the correspondences between words and the visual input \cite{SemanticAttention,GLIED,MARN}, SGN leverages partially decoded caption to discover the hierarchical structure of video by associating frames with phrases of the caption.

\textbf{Supervision for Attention.}
\label{subsec:supervision}
In video captioning, the attention mechanism is commonly used in the forms of temporal \cite{MARN,HLSTM-ATA,TDDF}, spatial \cite{MGSA,MAM,hRNN}, and regional \cite{zhang2019object,zhao2018video,wu2018interpretable}. 
In order to place a more precise attention mechanism, several methods have been proposed to provide explicit supervision by directly exploiting various human perception.
AC \cite{AC} employs a human-annotated binary attention mask to improve the correctness of spatial attention maps.
GVD \cite{GVD} explicitly links each noun phrase in a caption with a corresponding bounding box in a video frame.
As another type of human perception, human gaze information is used for recognizing important regions to look at in each video frame \cite{HumanGaze}.
These methods utilize the human perception obtained by manual annotations, which are expensive and not scalable to different datasets as they require considerable human efforts. 
However, the proposed CA loss makes our method take advantage of human perception through labor-free supervision and can be readily applied to other captioning datasets.

\section{Semantic Grouping Network}
\label{sec:method}

As shown in Fig. \ref{fig:architecture}, the Semantic Grouping Network (SGN) consists of four components:
(a) \textbf{Visual Encoder} takes a video and produces frame representations for each video frame, (b) \textbf{Phrase Encoder} takes the partially decoded caption and produces phrases consisting of a set of words in the caption, (c) \textbf{Semantic Grouping} filters out similar phrases and constructs semantic groups by aligning frames around the surviving phrases, and (d) \textbf{Decoder} exploits the semantic groups to predict the next word of the partially decoded caption.
For training, the \textbf{Contrastive Attention (CA) loss} is proposed to make the semantic groups have more coherent meanings within each group.
The details of each component will be described in the following subsections.

\subsection{Visual Encoder}
Given an input video $V$, $N$ frames $\{f_i\}_{i=1}^{N}$ and clips $\{c_i\}_{i=1}^{N}$ are uniformly sampled where each clip $c_i$ consists of consecutive frames around each sampled frame $f_i$. 
CNNs have demonstrated their strengths in encoding images and videos on various computer vision tasks such as classification \cite{kim2019edge,ju2020gapnet}, VQA \cite{kim2020modality,kim2019progressive}, and object detection \cite{chen2020memory,vu2019cascade}.
The appearance representations $\{\mathbf{v}_i^a\}_{i=1}^N$ and the motion representations $\{\mathbf{v}_i^m\}_{i=1}^N$ are extracted from a pre-trained 2D-CNN $\phi^a$ and 3D-CNN $\phi^m$, respectively. 
The two types of visual representations are then concatenated frame by frame to produce frame representations $\{\mathbf{v}_i\}_{i=1}^N$ as 
\begin{equation}
    \mathbf{v}_i = [ \mathbf{v}^a_i; \mathbf{v}^m_i ],
\end{equation}
where $\mathbf{v}^a_i = \phi^a(f_i)$, $\mathbf{v}^m_i = \phi^m(c_i)$, and $[\cdot; \cdot]$ denotes concatenation.

\subsection{Phrase Encoder}
A phrase is better than a word when identifying relevant frames.
There are words that do not have concrete and observable meanings when used alone, for example, function words like ``is" and ``the". 
In addition, a word may have insufficient meaning to specify related frames; for instance, it would be more explicit to associate frames containing ``man with glasses" rather than ``man" or ``glasses".
Therefore, a phrase rather than a word is used when performing a visual-textual alignment. 

To build phrases from the partially decoded caption, it is important to model the dependency among the words and see how they are related.
When generating the $t$-th word $w_t$ of the caption, we have a word representation matrix $W_t = [E[w_1] \cdots E[w_{t-1}]]^T \in \mathbb{R}^{(t-1) \times d_w}$ where $E$ denotes a word embedding matrix. 
Phrase Encoder $\phi^p$ takes the word representation matrix $W_t$ and produces a phrase representation matrix $P_t = [\mathbf{p}_{1,t} \cdots \mathbf{p}_{t-1, t}]^T \in \mathbb{R}^{(t-1) \times d_w}$ as
\begin{equation}
    P_t, A_t = \phi^p(W_t),
\end{equation}
where $A_t = [ \mathbf{a}_{1,t} \cdots \mathbf{a}_{t-1,t} ]^T \in \mathbb{R}^{(t-1) \times (t-1)}$ is a word attention matrix and $\mathbf{a}_{j,t} \in \mathbb{R}^{t-1}$ is the attention weights for the words $\{w_i\}_{i=1}^{t-1}$ used when constructing the phrase $p_{j,t}$.
For the Phrase Encoder $\phi^p$, the self-attention mechanism \cite{Transformer} is adopted as it is well known for modeling such intra-dependencies between words in a sentence. 
Although the phrases are constructed using the same set of words, the phrases that form semantic groups are learned to be discriminative by the Phrase Suppressor, which will be detailed in the following.

\subsection{Semantic Grouping}
A word phrase is the basis of a semantic group, which consists of the phrase and all the frames semantically linked to the phrase. The number of candidate phrases is the same as the number of words, and it turns out that many of the phrases generated by the Phrase Encoder are very similar. It would be better to filter out these phrases, and this is performed with a \textit{Phrase Suppressor}. Once a set of distinctive phrases is obtained, the \textit{Semantic Aligner} aligns the video frames to the surviving phrases.

\textbf{Phrase Suppressor.}
To leave only the distinctive phrases among all candidate phrases of $P_t$, Phrase Suppressor measures the degree of similarity between phrases and decides which ones to leave and which ones to discard. 
To do so, the similarities of all phrase pairs are measured by the outer product of the word attention matrix as $R_t = A_t (A_t)^T$ where $r_{i,j,t}$ measures how similar the two phrases $p_{i,t}$ and $p_{j,t}$ are.
A pair of two phrases $p_{i,t}$ and $p_{j,t}$ is considered to be similar if $r_{i,j,t}$ is larger than some fixed threshold $\tau$.
If so, the phrase that is more similar to other phrases is considered as the redundant phrase and discarded.
For example, if $r_{i,j,t} > \tau$ and $\sum_k r_{i,k,t} > \sum_k r_{j,k,t}$, the phrase $p_{i,t}$ is the redundant phrase and the phrase $p_{j,t}$ will survive. The procedure of phrase suppression are shown in Algorithm \ref{alg:phr-supp}.

\begin{algorithm}[t]
    \caption{Phrase Suppression.}
    \label{alg:phr-supp}
    \begin{algorithmic}[1]
        \Require{Phrases $\mathcal{P} = \{p_1,\cdots,p_K\}$, a word attention matrix $A$, and a threshold $\tau$} 
        \Ensure{A filtered set of phrases $\widehat{\mathcal{P}} = \{ \widehat{p}_1,\cdots,\widehat{p}_{K^\prime} \}$.}
        \Function{PhraseSuppressor}{$P$, $A$, $\tau$}
            \State {$\widehat{\mathcal{P}} \gets \mathcal{P}$}
            \State {$R \gets A A^T$}
            \For {$r_{i,j} \in \{r_{i,j} | r_{i,j} \in R, r_{i,j} > \tau\}$}
                \If {$\sum_k r_{i,k} > \sum_k r_{j,k}$}
                    \State{$\widehat{\mathcal{P}} \gets \widehat{\mathcal{P}} \setminus \{p_i\}$}
                \Else
                    \State{$\widehat{\mathcal{P}} \gets \widehat{\mathcal{P}} \setminus \{p_j\}$}
                \EndIf
            \EndFor
            \State \Return {$\widehat{\mathcal{P}}$}
        \EndFunction
    \end{algorithmic}
\end{algorithm}

After the phrase suppression, assume there are $M_t$ surviving phrases. Denote the surviving phrase representation matrix as $\widehat{P}_t = [ \widehat{\mathbf{p}}_{1,t} \cdots \widehat{\mathbf{p}}_{M_t,t} ]^T \in \mathbb{R}^{M_t \times d_w}$.
To this end, phrases are composed of different combinations of words, and the SGN is encouraged to form unique semantic groups.

\textbf{Semantic Aligner.}
For each pair of a phrase $\widehat{p}_{i,t}$ and a frame $f_j$, a score $\alpha_{i,j,t}$ is assigned based on the relevance between their representation vectors. 
As a common practice for measuring the relevance between two vectors, we obtain the relevance scores as
\begin{equation}
    \label{eqn:ca}
    \alpha_{i,j,t} \propto \mathbf{u}_s^T \sigma(U_s \widehat{\mathbf{p}}_{i,t} + H_s \mathbf{v}_j + \mathbf{b}_s),
\end{equation}
where $\mathbf{u}_s$, $U_s$, $H_s$, and $\mathbf{b}_s$ are learnable parameters, and $\sigma$ is an activation function such as hyperbolic tangent.

The relevance scores are normalized using softmax and then used to select which frames to be aligned with the phrase $\widehat{p}_{i,t}$ when computing the aligned frame representation $\mathbf{v}^p_{i,t}$ (see Equation \ref{eqn:aligned_frame_repr}). Finally, the representation for the semantic group $s_{i,t}$ around the phrase $\widehat{p}_{i,t}$ is obtained as
\begin{eqnarray}
    \label{eqn:aligned_frame_repr}
    \mathbf{v}^p_{i,t} &=& \sum_{j=1}^{N} \alpha_{i,j,t} \mathbf{v}_j, \\
    \mathbf{s}_{i,t} &=& [ \widehat{\mathbf{p}}_{i,t}; \mathbf{v}^p_{i,t} ].
\end{eqnarray}
By using the semantic groups $\{s_{i,t}\}_{i=1}^{M_t}$ as information units on behalf of the frames $\{f_j\}_{j=1}^N$ and words $\{w_i\}_{i=1}^{t-1}$, redundancy from adjacent frames is avoided and the decoder can exploit information units with more concrete meanings.

\subsection{Decoder}
Once semantic groups are constructed, a decoder extracts the necessary information for predicting the next word $w_t$. The decoder assigns a score to each semantic group that represents the usefulness in predicting the next word based on the correspondence with the previous decoder state $\mathbf{h}_{t-1}$ as
\begin{eqnarray}
    \beta_{i,t} &\propto& \mathbf{u}_d^T \sigma(U_d \mathbf{h}_{t-1} + H_d \mathbf{s}_{i,t} + \mathbf{b}_d), \\
    \mathbf{x}_t &=& \sum_{i=1}^{M_t} \beta_{i,t} \mathbf{s}_{i,t},
\end{eqnarray}
where $\mathbf{u}_d$, $U_d$, $H_d$, and $\mathbf{b}_d$ are learnable parameters.

Then, $\mathbf{x}_t$ is passed to an LSTM, and the probability distribution of the next word is generated by a fully connected layer followed by a softmax layer as
\begin{eqnarray}
    &&\mathbf{h}_{t} = \textup{LSTM}([\mathbf{x}_{t}; E[w_{t-1}]], \mathbf{h}_{t-1}), \\
    &&p(w_t | V, w_1, \cdots, w_{t-1}) = \textup{softmax}(U_h \mathbf{h}_{t} + \mathbf{b}_h),
\end{eqnarray}
where $U^h$ and $\mathbf{b}^h$ are learnable parameters.
Our decoder is the same with that of the typical video captioning method \cite{TA}, except the term ``temporal attention" is replaced with ``semantic attention" since the targets of the attention are not the frames, but semantic groups.

\subsection{Training}
\label{sec:loss}
One of the most crucial points in training the SGN is to induce the generation of \textit{distinctive} and \textit{coherent} semantic groups.
For distinctive semantic groups, the Phrase Suppressor filtered out redundant phrases.
For coherent semantic groups, in addition to the typical cross-entropy loss $\mathcal{L}_{ce}$ for caption generation, a Contrastive Attention loss $\mathcal{L}_{ca}$ is introduced.
Given a video $V$ and its ground-truth caption $Y=[y_1,\cdots,y_T]$ from a training dataset $\mathcal{D}$, the loss function $\mathcal{L}$ is formulated as
\begin{equation}
    \mathcal{L} = \mathcal{L}_{ce} + \lambda \mathcal{L}_{ca}.
\end{equation}

\textbf{Cross-Entropy Loss.}
Cross-entropy loss is defined as the negative log-likelihood to generate the correct caption:
\begin{equation}
    \mathcal{L}_{ce} = \sum_{(V, Y) \in \mathcal{D}} \sum_t \left( -\log p(y_t | V, y_1, \cdots, y_{t-1}) \right).
\end{equation}


\textbf{Contrastive Attention Loss.}
%
The semantic group should only contain the frames that are highly related to their phrase to ensure a semantic group to have a coherent meaning across its members.
To this end, a \textit{negative video} of the input video is sampled, and its frames, referred to as \textit{negative frames}, are provided as erroneous candidates for the Semantic Aligner. 
In order to ensure that it is completely irrelevant to the input video, the negative video is randomly sampled from a set of videos whose caption does not overlap with that of the input video; two captions are said to be overlapped if a word excluding stopwords (e.g., ``a", ``the") is included in both captions.
The positive relevance score $\alpha^{pos}_{i,j,t}$ between a phrase $\widehat{p}_i$ and an input frame $f_j$, and the negative relevance score $\alpha^{neg}_{i,j,t}$ between a phrase $\widehat{p}_i$ and a negative frame $f^{neg}_j$ are obtained by following Equation \ref{eqn:ca}.
Then, the relevance scores are normalized by applying softmax given both positive and negative relevance scores, and $p_{ca}(s_{i,t}) = \sum_{j=1}^{N} \alpha^{pos}_{i,j,t}$ represents the probability that the semantic group $s_{i,t}$ will not contain any negative frames.
The $p_{ca}(s_{i,t})$ increases with an increase in positive relevance scores relative to the negative relevance scores, which is why the loss is referred to as ``Contrastive Attention loss”.
The CA loss is formulated as
\begin{equation}
    \mathcal{L}_{ca} = \sum_{(V, Y) \in \mathcal{D}} \sum_t \sum_i^{M_t} \left( -\textup{log}~p_{ca}(s_{i,t}) \right).
\end{equation}

\section{Experiments}

\begin{table*}[t]
    \centering
    \begin{tabular}{l|c|cccccccc}
        \hline
        \hline
                                                       &          & \multicolumn{4}{c}{MSVD}                                    & \multicolumn{4}{c}{MSR-VTT}                                   \\
        Model                                          & Detector & ~B@4~       & ~~C~~         & ~~M~~         & ~~R~~         & ~B@4~         & ~~C~~       & ~~M~~         & ~~R~~           \\
        \hline
        TA (G) \cite{TA}                               & \xmark & 41.9          & 51.7          & 29.6          & -             & -             & -             & -             & -             \\
        HRNE (G) \cite{HRNE}                           & \xmark & 43.8          & -             & 33.1          & -             & -             & -             & -             & -             \\
        $\textup{MGSA}^\dagger$ (G+O) \cite{MGSA}      & \xmark & 49.5          & 74.2          & 32.2          & -             & 39.9          & 45.0          & 26.3          & -             \\
        MAM (V) \cite{MAM}                             & \xmark & 41.3          & 53.9          & 32.2          & 68.8          & -             & -             & -             & -             \\
        h-RNN (V) \cite{hRNN}                          & \xmark & 44.3          & 62.1          & 31.1          & -             & -             & -             & -             & -             \\
        M3 (V) \cite{M3}                               & \xmark & 49.6          & -             & 30.1          & -             & 35.0          & -             & 24.6          & -             \\
        BAE (R50+C) (Baraldi et al. 2017)              & \xmark & 42.5          & 63.5          & 32.4          & -             & -             & -             & -             & -             \\
        hLSTMat (R152) \cite{HLSTM-ATA}                & \xmark & \textbf{53.0} & 73.8          & 33.6          & -             & 38.3          & -             & 26.3          & -             \\
        PickNet   (R152) \cite{PickNet}                & \xmark & 52.3          & 76.5          & 33.3          & 69.6          & 39.4          & 42.3          & 27.3          & 59.7          \\
        $\textup{MARN}^\dagger$ (R101+RN) \cite{MARN}  & \xmark & 48.6          & 92.2          & 35.1          & 71.9          & 40.4          & 47.1          & 28.1          & 60.7          \\
        \hline
        OA-BTG (R200) \cite{OA-BTG}                    & \cmark & 56.9          & 90.6          & 36.2          & -             & 41.4          & 46.9          & 28.2          & -             \\
        STG-KD (R101+RN) \cite{STG-KD}                 & \cmark & 52.2          & 93.0          & 36.9          & 73.9          & 40.5          & 47.1          & 28.3          & 60.9          \\
        $\textup{SAAT}^\dagger$ (IRV2+C3D) (Zheng et al. 2020) & \cmark & 46.5          & 81.0          & 33.5          & 69.4          & 40.5          & 49.1          & 28.2          & 60.9  \\
        ORG-TRL (IRV2+C3D) \cite{ORG-TRL}              & \cmark & 54.3          & 95.2          & 36.4          & 73.9          & 43.6          & 50.9          & 28.8          & 62.1          \\
        \hline
        SGN (G)                                        & \xmark & 46.3          & 73.2          & 32.1          & 67.3          & 37.3          & 41.2          & 26.8          & 58.2          \\
        SGN (V)                                        & \xmark & 47.7          & 74.9          & 33.1          & 69.0          & 37.8          & 41.9          & 27.0          & 58.3          \\
        SGN (R152)                                     & \xmark & 48.2          & 84.6          & 34.2          & 69.8          & 39.6          & 45.2          & 27.6          & 59.6          \\
        SGN (R101+RN)                                  & \xmark & 52.8          & \textbf{94.3} & \textbf{35.5} & \textbf{72.9} & \textbf{40.8} & \textbf{49.5} & \textbf{28.3} & \textbf{60.8} \\
        \hline
        \hline
    \end{tabular}
    \caption{Quantitative results on MSVD and MSR-VTT datasets. G, V, R, C, RN, and O denote GoogLeNet, VGGNet-19, ResNet, C3D, 3D-ResNext-101, and Optical Flow, respectively. B@4, C, M, and R denote BLEU@4, CIDEr-D, METEOR, and ROUGE\_L, respectively. Methods with a dagger ($\dagger$) utilize video categories as auxiliary data on the MSR-VTT dataset.}
    \label{table:quantitative}
\end{table*}

\subsection{Experimental Setup}

Various experiments are conducted to show the effectiveness of SGN using the two most popular benchmark datasets.

\textbf{MSVD.}
Microsoft Video Description (MSVD) dataset \cite{MSVD}, also known as YoutubeClips, contains $1970$ YouTube videos whose average length is about $10$ seconds. 
Each video is described with $40$ English sentences written by Amazon Mechanical Turks. 
For a fair comparison, the dataset is divided into a training set of $1200$ videos, a validation set of $100$ videos, and a test set of $670$ videos by following the official split \cite{TA}.

\textbf{MSR-VTT.}
MSR Video-to-Text (MSR-VTT) dataset \cite{MSR-VTT} is a large-scale benchmark dataset. It contains $10000$ videos whose average length is about $20$ seconds, and each video is annotated with $20$ English captions and a category tag. Following Xu et al. \cite{MSR-VTT}, the dataset is divided into a training set of $6513$ videos, a validation set of $497$ videos, and a test set of $2990$ videos.

\textbf{Implementation Details.} 
We uniformly sample $N=30$ frames and clips from each video. 
As video captioning performances depend on backbone CNNs, various pre-trained CNNs including GoogLeNet \cite{GoogLeNet}, VGGNet \cite{VGGNet}, ResNet \cite{ResNet}, and 3D-ResNext \cite{3DResNext} are employed as a Visual Encoder to fairly compare SGN with state-of-the-art methods.
The word embedding matrix is initialized using GloVe \cite{GloVe} and jointly trained with the whole architecture.
Before the first word ($w_1$) is generated, $<$SOS$>$ is used as the partially decoded caption (i.e., $w_0 = <$SOS$>$) and then ignored thereafter.
$\tau$ and $\lambda$ are set to $0.2$ and $0.16$ as a result of 5-fold cross-validation for the values of $[0.1, 0.2, 0.3]$ and $[0.01, 0.04, 0.16, 0.64]$, respectively.
Beam search with a size of $5$ is used for generating the final captions.
BLEU@4 \cite{BLEU}, CIDEr-D \cite{CIDEr}, METEOR \cite{METEOR}, and ROUGE\_L \cite{ROUGE} are used for evaluation, and the scores are computed using the official codes from Microsoft COCO evaluation server \cite{COCO-eval-server}.

\subsection{Quantitative Results}
We compare the performance of the SGN with that of state-of-the-art methods based on three different approaches:
(1) Encoding Video into Information Units: HRNE \cite{HRNE}, BAE \cite{BAE}, and PickNet \cite{PickNet}, which are described eariler.
(2) Multi-modal Reasoning: M3 \cite{M3} and MARN \cite{MARN}, which are described eariler.
(3) Applying Temporal/Spatial Attention: TA \cite{TA} was the first to introduce temporal attention for exploiting the temporal structure of the video. h-RNN \cite{hRNN} and MAM \cite{MAM} exploit both temporal and spatial attention to focus on the most correlated frames as well as salient regions. hLSTMat \cite{HLSTM-ATA} employs a hierarchical LSTM to adjust temporal attention weights and decides which modality to use for predicting the next word. MGSA \cite{MGSA} utilizes optical flow as supervision to learn spatial attention maps.
We did not compare the SGN with methods that utilize detectors. 
Also, we follow the standard practice to not compare to methods based on RL.

The results are shown in Table \ref{table:quantitative}. 
For both datasets, SGN outperforms most state-of-the-art methods with various backbone CNNs, especially on the CIDEr-D metric.
Note that CIDEr-D is specifically designed for the captioning task and is known to be more consistent with human judgment than the others. 
SGN outperforms the runner-up methods by large margins of 2.1\%p and 2.4\%p in a CIDEr-D score on MSVD and MSR-VTT datasets, respectively. 
With the GoogLeNet feature, SGN could not beat the MGSA \cite{MGSA}; although MGSA additionally utilizes Optical Flow and video category tag for MSR-VTT, it implies exploiting spatial structure would be helpful, which remains our future works.

\begin{table}[h]
    \centering
    \begin{tabular}{ccc|cccc}
    \hline
    \hline
    SA     & PS     & CA      & B@4           & C             & M             & R \\ 
    \hline
    \xmark & \xmark & \xmark  & 37.9	      & 42.2          & 26.1          & 58.8 \\ 
    \cmark & \xmark & \xmark  & 40.0          & 47.9          & 27.8          & 60.2 \\ 
    \cmark & \cmark & \xmark  & 40.2          & 48.5          & 27.9          & 60.2 \\ 
    \cmark & \cmark & \cmark  & \textbf{40.8} & \textbf{49.5} & \textbf{28.3} & \textbf{60.8} \\
    \hline
    \hline
    \end{tabular}
    \caption{Performance on the MSR-VTT dataset with different add-on components. SA, PS, and CA denote a Semantic Aligner (including Phrase Encoder), a Phrase Suppressor, and a Contrastive Attention loss, respectively.}
    \label{table:ablation-component}
\end{table}

\textbf{Ablation Study.}
In order to evaluate the effectiveness of the proposed components in SGN, Table \ref{table:ablation-component} shows the ablation test results.
SA enables SGN to represent video as semantic information units and understand it based on them. PS eliminates similar phrases and prevents the formation of redundant semantic groups. CA facilitates accurate alignment between word phrases and video frames in constructing semantic groups.
If SGN does not have any proposed components (1st row), it becomes TA \cite{TA}; The TA and SGN differ in that the former assigns the attention weights in each frame, and the latter assigns those per semantic group. 
It is observed that as we enhance the functionality of the baseline TA with a semantic aligner, phrase suppressor, and contrastive attention loss, the performance generally rises with respect to all of the measures.
The significant increase in performance when semantic aligner is introduced (1st to 2nd row) means that encoding the video by splitting it into units of information based on the meaning of the partially decoded caption helps the model understand the comprehensive caption context.

To confirm the claim that phrases are better than words when forming semantic groups, the performance of SGN without the Phrase Encoder and Phrase Suppressor is shown in Table \ref{table:ablation-phrase}.
It is observed that SGN significantly outperforms its counterpart in terms of all metrics, which implies that it is hard to form semantic groups based on a single word.

\begin{table}[h]
    \centering
    \begin{tabular}{c|cccc}
    \hline
    \hline
        Model               & B@4           & C             & M             & R             \\ 
    \hline
        SGN (group by word) & 39.9          & 46.6          & 27.0          & 59.3          \\ 
        SGN                 & \textbf{40.8} & \textbf{49.5} & \textbf{28.3} & \textbf{60.8} \\ 
    \hline
    \hline
    \end{tabular}
    \caption{Performance of SGN that forms semantic groups using phrases (default) and words on the MSR-VTT dataset.}
    \label{table:ablation-phrase}
\end{table}

\textbf{Inference Speed.}
On a single Titan V GPU with 12GB of memory, we measured the inference speed of two methods, SGN and TA \cite{TA} (see Table \ref{table:inference-speed}).
The CNN features are pre-extracted, and beam search is not used in this experiment.
TA is similar to SGN in that it directly applies temporal attention to video frames without a grouping process.
The time complexity of SGN to predict the $t$-th word is $O((t-1)^2 + (t-1)^2 + Nt) \approx O(Nt)$ ($\because t < N$), where $N$ is the number of sampled frames, while that of TA is $O(N)$. Measured inference speeds of TA and SGN were respectively 865 and 657 videos per second on MSVD, and 268 and 203 videos per second on MSR-VTT.
SGN's repeated grouping process reduces the inference speed of about 25\%, but it still attains latency of less than 10ms and results in a significant boost in captioning performance.

\begin{table}[h]
    \centering
    \begin{tabular}{c|c|cc}
    \hline
    \hline
        Model & Time complexity & MSVD & MSR-VTT \\ 
    \hline
        TA    & $O(N)$          & 865  & 268     \\
        SGN   & $O(Nt)$         & 657  & 203     \\
    \hline
    \hline
    \end{tabular}
    \caption{Inference speed of SGN and TA \cite{TA} in terms of the number of decoded videos per second.}
    \label{table:inference-speed}
\end{table}

\subsection{Qualitative Results}
\begin{figure}[t]
    \begin{subfigure}{0.5\textwidth}
      \centering
      \includegraphics[width=\linewidth]{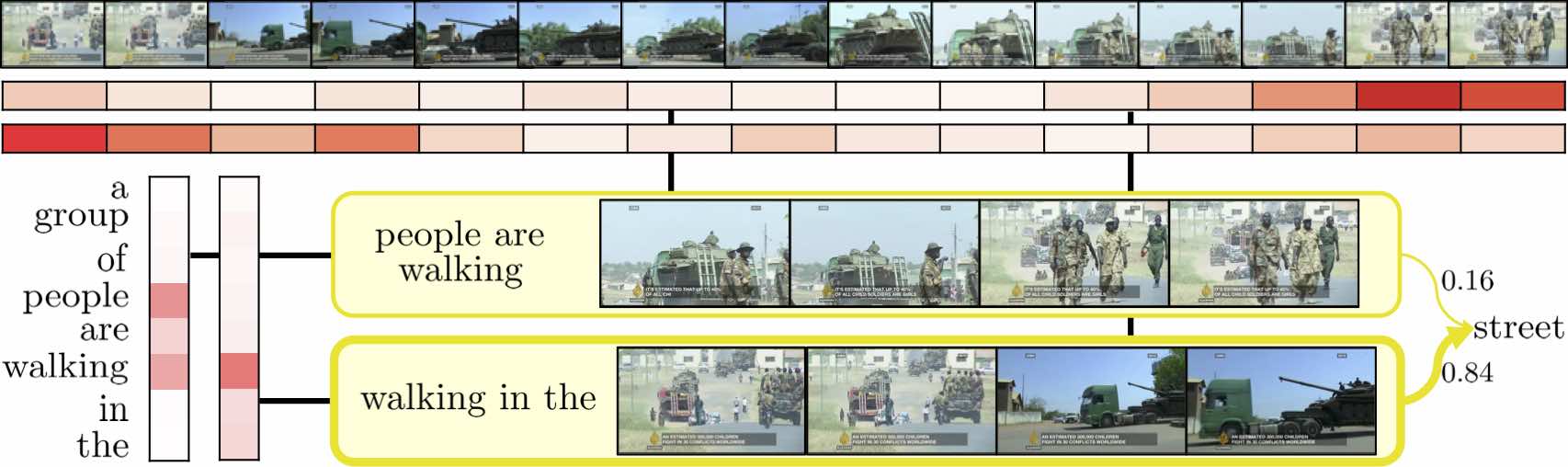}
      \caption{\small }
      \label{fig:qualitative_semantic_groups_a}
    \end{subfigure}
    \\
    \begin{subfigure}{0.5\textwidth}
      \centering
      \includegraphics[width=\linewidth]{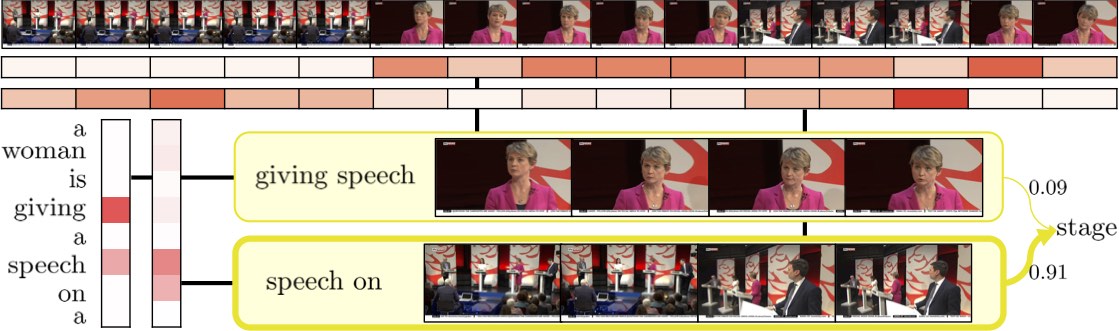}
      \caption{\small }
      \label{fig:qualitative_semantic_groups_b}
    \end{subfigure}
    \caption{Illustrations of showing how semantic groups are formed and exploited. Red color indicates the magnitude of an attention weight assigned to a word (left) or frame (top).}
    \label{fig:qualitative_semantic_groups}
\end{figure}

\textbf{Semantic Groups.}
To see how the semantic groups are formed and exploited by SGN, Figure \ref{fig:qualitative_semantic_groups} provides two examples.
In Figure \ref{fig:qualitative_semantic_groups_a}, the two phrases ``people are walking`` and ``walking in the" are constructed using words in the partially decoded caption ``a group of people are walking in the".
One semantic group is formed by gathering frames that show people (soldiers) walking, and the other is formed by gathering the frames that capture where walking is possible. 
The latter semantic group is exploited more in predicting the next word ``street".
Similarly, in Figure \ref{fig:qualitative_semantic_groups_b}, two semantic groups are formed based on phrases ``giving speech" and ``speech on", and the latter is exploited to predict the next word ``stage".
These visualizations show that the SGN has the potential to generate discriminating phrases and accurately associate the frames to the phrases. 
The semantic groups defined by the phrases are explainable and can be correctly exploited to generate the next word in the caption.

\textbf{Caption Results.}
\begin{figure}[t]
    \centering
    \begin{subfigure}{.47\textwidth}
      \centering
      \includegraphics[width=\linewidth]{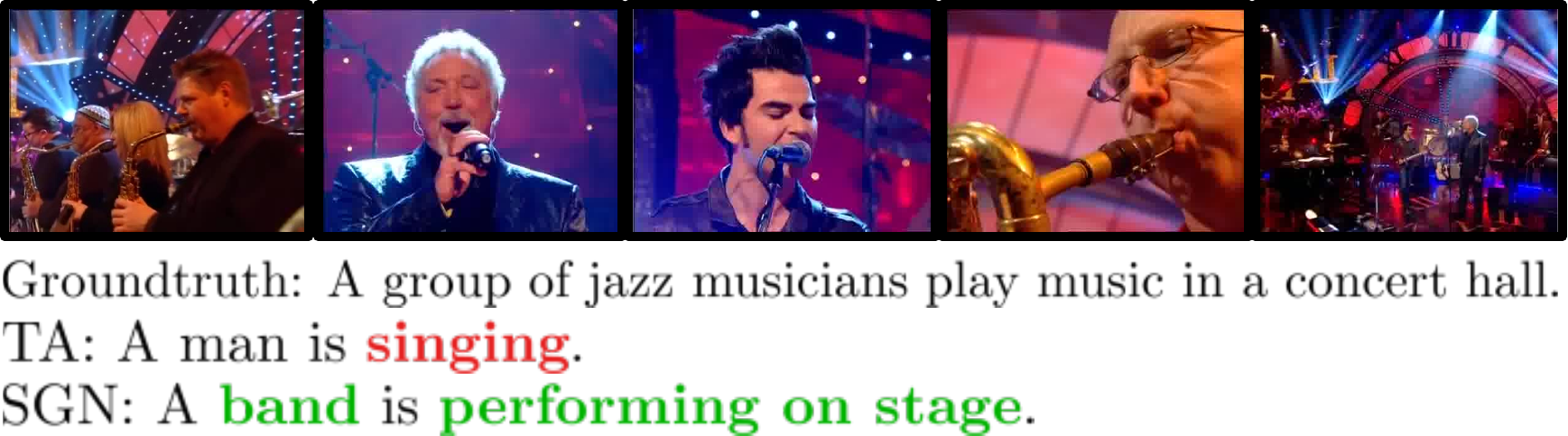}
      \caption{\small }
      \label{fig:caption_results_a}
    \end{subfigure}%
    \\
    \begin{subfigure}{.47\textwidth}
      \centering
      \includegraphics[width=\linewidth]{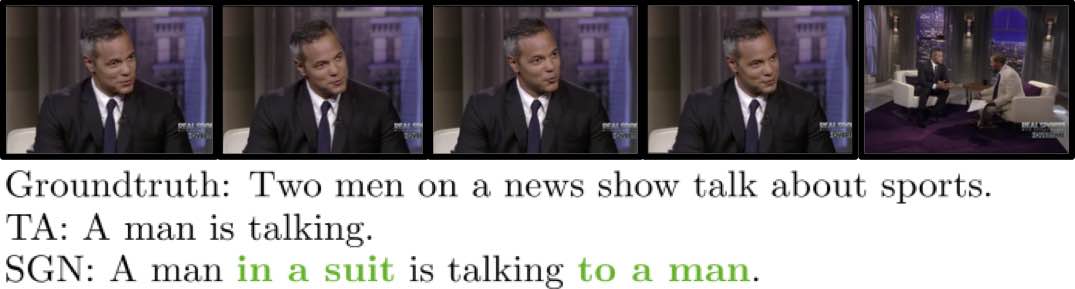}
      \caption{\small }
      \label{fig:caption_results_b}
    \end{subfigure}
    \\
    \begin{subfigure}{.47\textwidth}
      \centering
      \includegraphics[width=\linewidth]{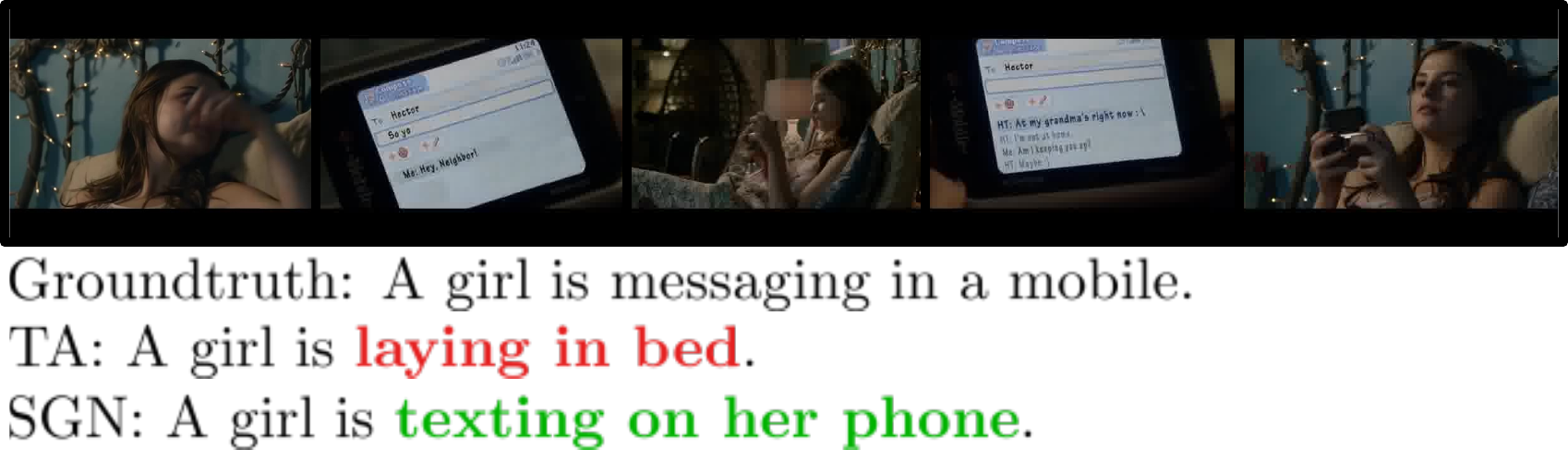}
      \caption{\small }
      \label{fig:caption_results_c}
    \end{subfigure}%
    \\
    \begin{subfigure}{.47\textwidth}
      \centering
      \includegraphics[width=\linewidth]{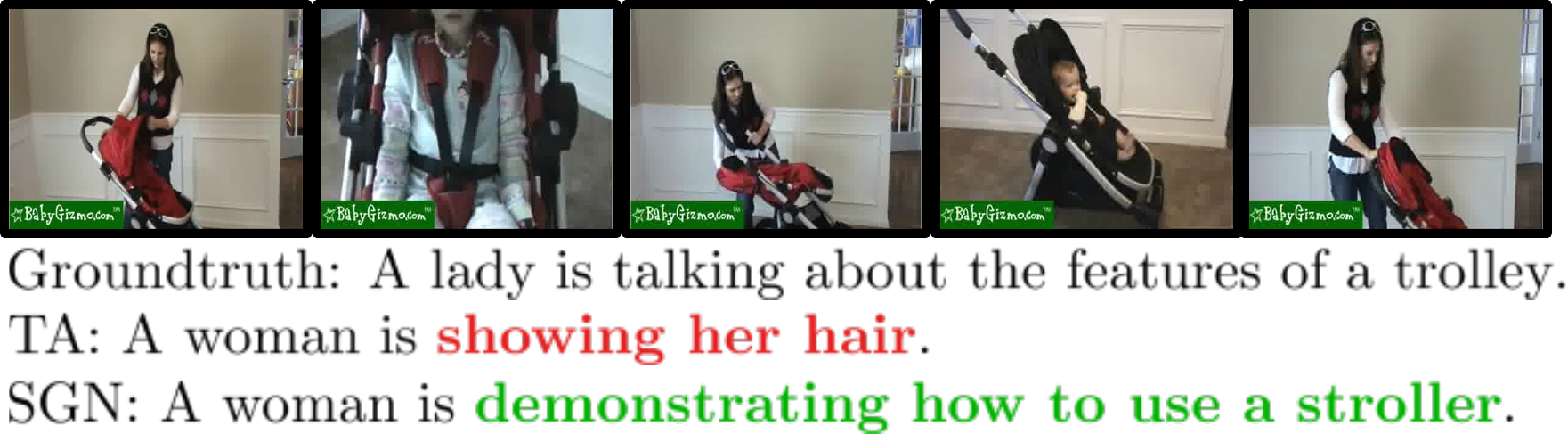}
      \caption{\small }
      \label{fig:caption_results_d}
    \end{subfigure}
    \caption{Illustrations of captions generated by SGN and TA.}
    \label{fig:caption_results}
\end{figure}
Fig. \ref{fig:caption_results} shows examples of caption generated by SGN and TA \cite{TA} - the same pre-trained CNN features are used.
SGN is able to better identify the subject responsible for the action performed in the lengthy video scene. For example, SGN chose to predict ``a band is performing” rather than ``a man is singing” (Figure \ref{fig:caption_results_a}), and it provides a more detailed description of the video (Figure \ref{fig:caption_results_b}).
Overall, SGN seems to understand the context better than TA as shown in Figure \ref{fig:caption_results_c} and \ref{fig:caption_results_d}.

\section{Conclusion}
In this paper, we propose a Semantic Grouping Network (SGN) for video captioning which has a comprehensive understanding of captioning context by encoding a video into semantic groups consisting of phrases of partially decoded caption and related frames.
In contrast to the prior methods, the continuous feedback from decoded words allows SGN to dynamically update the video representation that adapts to the partially decoded caption.
The Contrastive Attention loss provides efficient supervision for correct visual-textual alignment within a semantic group without requiring any manual annotations.
The constructed semantic groups are explainable as each of them has a distinct meaning and has coherent semantics shared across its members, and are exploited to predict the next word.
The SGN achieves state-of-the-art performances by outperforming runner-up methods by large margins of 2.1\%p and 2.4\%p in terms of the CIDEr-D metric on MSVD and MSR-VTT datasets, respectively.

\bibliography{refs}

\begin{thebibliography}{42}
\providecommand{\natexlab}[1]{#1}
\providecommand{\url}[1]{\texttt{#1}}
\providecommand{\urlprefix}{URL }
\expandafter\ifx\csname urlstyle\endcsname\relax
  \providecommand{\doi}[1]{doi:\discretionary{}{}{}#1}\else
  \providecommand{\doi}{doi:\discretionary{}{}{}\begingroup
  \urlstyle{rm}\Url}\fi

\bibitem[{Banerjee and Lavie(2005)}]{METEOR}
Banerjee, S.; and Lavie, A. 2005.
\newblock METEOR: An automatic metric for MT evaluation with improved
  correlation with human judgments.
\newblock In \emph{Proceedings of the ACL Workshop on Intrinsic and Extrinsic
  Evaluation Measures for Machine Translation and/or Summarization}.

\bibitem[{Baraldi, Grana, and Cucchiara(2017)}]{BAE}
Baraldi, L.; Grana, C.; and Cucchiara, R. 2017.
\newblock Hierarchical boundary-aware neural encoder for video captioning.
\newblock In \emph{Proceedings of the IEEE Conference on Computer Vision and
  Pattern Recognition}.

\bibitem[{Chen and Dolan(2011)}]{MSVD}
Chen, D.~L.; and Dolan, W.~B. 2011.
\newblock Collecting highly parallel data for paraphrase evaluation.
\newblock In \emph{Proceedings of the 49th Annual Meeting of the Association
  for Computational Linguistics}.

\bibitem[{Chen and Jiang(2019)}]{MGSA}
Chen, S.; and Jiang, Y.-G. 2019.
\newblock Motion Guided Spatial Attention for Video Captioning.
\newblock In \emph{Proceedings of AAAI Conference on Artificial Intelligence}.

\bibitem[{Chen et~al.(2015)Chen, Fang, Lin, Vedantam, Gupta, Doll{\'a}r, and
  Zitnick}]{COCO-eval-server}
Chen, X.; Fang, H.; Lin, T.-Y.; Vedantam, R.; Gupta, S.; Doll{\'a}r, P.; and
  Zitnick, C.~L. 2015.
\newblock Microsoft coco captions: Data collection and evaluation server.
\newblock \emph{arXiv preprint arXiv:1504.00325} .

\bibitem[{Chen et~al.(2020)Chen, Cao, Hu, and Wang}]{chen2020memory}
Chen, Y.; Cao, Y.; Hu, H.; and Wang, L. 2020.
\newblock Memory Enhanced Global-Local Aggregation for Video Object Detection.
\newblock In \emph{Proceedings of the IEEE/CVF Conference on Computer Vision
  and Pattern Recognition}.

\bibitem[{Chen et~al.(2018)Chen, Wang, Zhang, and Huang}]{PickNet}
Chen, Y.; Wang, S.; Zhang, W.; and Huang, Q. 2018.
\newblock Less is more: Picking informative frames for video captioning.
\newblock In \emph{Proceedings of the European Conference on Computer Vision}.

\bibitem[{Hara, Kataoka, and Satoh(2018)}]{3DResNext}
Hara, K.; Kataoka, H.; and Satoh, Y. 2018.
\newblock Can spatiotemporal 3d cnns retrace the history of 2d cnns and
  imagenet?
\newblock In \emph{Proceedings of the IEEE Conference on Computer Vision and
  Pattern Recognition}.

\bibitem[{He et~al.(2016)He, Zhang, Ren, and Sun}]{ResNet}
He, K.; Zhang, X.; Ren, S.; and Sun, J. 2016.
\newblock Deep residual learning for image recognition.
\newblock In \emph{Proceedings of the IEEE Conference on Computer Vision and
  Pattern Recognition}.

\bibitem[{Ju et~al.(2020)Ju, Ryu, Moon, and Yoo}]{ju2020gapnet}
Ju, M.; Ryu, H.; Moon, S.; and Yoo, C.~D. 2020.
\newblock GAPNet: Generic-Attribute-Pose Network For Fine-Grained Visual
  Categorization Using Multi-Attribute Attention Module.
\newblock In \emph{2020 IEEE International Conference on Image Processing}.

\bibitem[{Kim et~al.(2019{\natexlab{a}})Kim, Kim, Kim, and Yoo}]{kim2019edge}
Kim, J.; Kim, T.; Kim, S.; and Yoo, C.~D. 2019{\natexlab{a}}.
\newblock Edge-labeling graph neural network for few-shot learning.
\newblock In \emph{Proceedings of the IEEE Conference on Computer Vision and
  Pattern Recognition}.

\bibitem[{Kim et~al.(2019{\natexlab{b}})Kim, Ma, Kim, Kim, and
  Yoo}]{kim2019progressive}
Kim, J.; Ma, M.; Kim, K.; Kim, S.; and Yoo, C.~D. 2019{\natexlab{b}}.
\newblock Progressive attention memory network for movie story question
  answering.
\newblock In \emph{Proceedings of the IEEE Conference on Computer Vision and
  Pattern Recognition}.

\bibitem[{Kim et~al.(2020)Kim, Ma, Pham, Kim, and Yoo}]{kim2020modality}
Kim, J.; Ma, M.; Pham, T.; Kim, K.; and Yoo, C.~D. 2020.
\newblock Modality Shifting Attention Network for Multi-Modal Video Question
  Answering.
\newblock In \emph{Proceedings of the IEEE/CVF Conference on Computer Vision
  and Pattern Recognition}.

\bibitem[{Li et~al.(2017)Li, Zhao, Lu et~al.}]{MAM}
Li, X.; Zhao, B.; Lu, X.; et~al. 2017.
\newblock MAM-RNN: Multi-level Attention Model Based RNN for Video Captioning.
\newblock In \emph{Proceedings of the 26th International Joint Conference on
  Artificial Intelligence}.

\bibitem[{Lin(2004)}]{ROUGE}
Lin, C.-Y. 2004.
\newblock Rouge: A package for automatic evaluation of summaries.
\newblock In \emph{Text summarization branches out}.

\bibitem[{Liu et~al.(2017)Liu, Mao, Sha, and Yuille}]{AC}
Liu, C.; Mao, J.; Sha, F.; and Yuille, A. 2017.
\newblock Attention correctness in neural image captioning.
\newblock In \emph{Proceedings of AAAI Conference on Artificial Intelligence}.

\bibitem[{Liu et~al.(2019)Liu, Ren, Liu, Lei, and Sun}]{GLIED}
Liu, F.; Ren, X.; Liu, Y.; Lei, K.; and Sun, X. 2019.
\newblock Exploring and distilling cross-modal information for image
  captioning.
\newblock In \emph{Proceedings of the 28th International Joint Conference on
  Artificial Intelligence}.

\bibitem[{Pan et~al.(2020)Pan, Cai, Huang, Lee, Gaidon, Adeli, and
  Niebles}]{STG-KD}
Pan, B.; Cai, H.; Huang, D.-A.; Lee, K.-H.; Gaidon, A.; Adeli, E.; and Niebles,
  J.~C. 2020.
\newblock Spatio-Temporal Graph for Video Captioning With Knowledge
  Distillation.
\newblock In \emph{Proceedings of the IEEE/CVF Conference on Computer Vision
  and Pattern Recognition}.

\bibitem[{Pan et~al.(2016)Pan, Xu, Yang, Wu, and Zhuang}]{HRNE}
Pan, P.; Xu, Z.; Yang, Y.; Wu, F.; and Zhuang, Y. 2016.
\newblock Hierarchical recurrent neural encoder for video representation with
  application to captioning.
\newblock In \emph{Proceedings of the IEEE Conference on Computer Vision and
  Pattern Recognition}.

\bibitem[{Papineni et~al.(2002)Papineni, Roukos, Ward, and Zhu}]{BLEU}
Papineni, K.; Roukos, S.; Ward, T.; and Zhu, W.-J. 2002.
\newblock BLEU: a method for automatic evaluation of machine translation.
\newblock In \emph{Proceedings of the 40th Annual Meeting on Association for
  Computational Linguistics}.

\bibitem[{Pei et~al.(2019)Pei, Zhang, Wang, Ke, Shen, and Tai}]{MARN}
Pei, W.; Zhang, J.; Wang, X.; Ke, L.; Shen, X.; and Tai, Y.-W. 2019.
\newblock Memory-Attended Recurrent Network for Video Captioning.
\newblock In \emph{Proceedings of the IEEE Conference on Computer Vision and
  Pattern Recognition}.

\bibitem[{Pennington, Socher, and Manning(2014)}]{GloVe}
Pennington, J.; Socher, R.; and Manning, C. 2014.
\newblock Glove: Global vectors for word representation.
\newblock In \emph{Proceedings of the 2014 Conference on Empirical Methods in
  Natural Language Processing}.

\bibitem[{Simonyan and Zisserman(2015)}]{VGGNet}
Simonyan, K.; and Zisserman, A. 2015.
\newblock Very Deep Convolutional Networks for Large-Scale Image Recognition.
\newblock In \emph{International Conference on Learning Representations}.

\bibitem[{Song et~al.(2017)Song, Gao, Guo, Liu, Zhang, and Shen}]{HLSTM-ATA}
Song, J.; Gao, L.; Guo, Z.; Liu, W.; Zhang, D.; and Shen, H.-T. 2017.
\newblock Hierarchical {LSTM} with Adjusted Temporal Attention for Video
  Captioning.
\newblock In \emph{Proceedings of the 26th International Joint Conference on
  Artificial Intelligence}.

\bibitem[{Szegedy et~al.(2015)Szegedy, Liu, Jia, Sermanet, Reed, Anguelov,
  Erhan, Vanhoucke, and Rabinovich}]{GoogLeNet}
Szegedy, C.; Liu, W.; Jia, Y.; Sermanet, P.; Reed, S.; Anguelov, D.; Erhan, D.;
  Vanhoucke, V.; and Rabinovich, A. 2015.
\newblock Going deeper with convolutions.
\newblock In \emph{Proceedings of the IEEE Conference on Computer Vision and
  Pattern Recognition}.

\bibitem[{Vaswani et~al.(2017)Vaswani, Shazeer, Parmar, Uszkoreit, Jones,
  Gomez, Kaiser, and Polosukhin}]{Transformer}
Vaswani, A.; Shazeer, N.; Parmar, N.; Uszkoreit, J.; Jones, L.; Gomez, A.~N.;
  Kaiser, {\L}.; and Polosukhin, I. 2017.
\newblock Attention is all you need.
\newblock In \emph{Advances in Neural Information Processing Systems}.

\bibitem[{Vedantam, Lawrence~Zitnick, and Parikh(2015)}]{CIDEr}
Vedantam, R.; Lawrence~Zitnick, C.; and Parikh, D. 2015.
\newblock Cider: Consensus-based image description evaluation.
\newblock In \emph{Proceedings of the IEEE Conference on Computer Vision and
  Pattern Recognition}.

\bibitem[{Vu et~al.(2019)Vu, Jang, Pham, and Yoo}]{vu2019cascade}
Vu, T.; Jang, H.; Pham, T.~X.; and Yoo, C. 2019.
\newblock Cascade RPN: Delving into High-Quality Region Proposal Network with
  Adaptive Convolution.
\newblock In \emph{Advances in Neural Information Processing Systems}.

\bibitem[{Wang et~al.(2018)Wang, Wang, Huang, Wang, and Tan}]{M3}
Wang, J.; Wang, W.; Huang, Y.; Wang, L.; and Tan, T. 2018.
\newblock M3: Multimodal memory modelling for video captioning.
\newblock In \emph{Proceedings of the IEEE Conference on Computer Vision and
  Pattern Recognition}.

\bibitem[{Wu et~al.(2018)Wu, Li, Cao, Ji, and Lin}]{wu2018interpretable}
Wu, X.; Li, G.; Cao, Q.; Ji, Q.; and Lin, L. 2018.
\newblock Interpretable video captioning via trajectory structured
  localization.
\newblock In \emph{Proceedings of the IEEE Conference on Computer Vision and
  Pattern Recognition}.

\bibitem[{Xu et~al.(2016)Xu, Mei, Yao, and Rui}]{MSR-VTT}
Xu, J.; Mei, T.; Yao, T.; and Rui, Y. 2016.
\newblock Msr-vtt: A large video description dataset for bridging video and
  language.
\newblock In \emph{Proceedings of the IEEE Conference on Computer Vision and
  Pattern Recognition}.

\bibitem[{Yao et~al.(2015)Yao, Torabi, Cho, Ballas, Pal, Larochelle, and
  Courville}]{TA}
Yao, L.; Torabi, A.; Cho, K.; Ballas, N.; Pal, C.; Larochelle, H.; and
  Courville, A. 2015.
\newblock Describing videos by exploiting temporal structure.
\newblock In \emph{Proceedings of the IEEE International Conference on Computer
  Vision}.

\bibitem[{You et~al.(2016)You, Jin, Wang, Fang, and Luo}]{SemanticAttention}
You, Q.; Jin, H.; Wang, Z.; Fang, C.; and Luo, J. 2016.
\newblock Image captioning with semantic attention.
\newblock In \emph{Proceedings of the IEEE Conference on Computer Vision and
  Pattern Recognition}.

\bibitem[{Yu et~al.(2016)Yu, Wang, Huang, Yang, and Xu}]{hRNN}
Yu, H.; Wang, J.; Huang, Z.; Yang, Y.; and Xu, W. 2016.
\newblock Video paragraph captioning using hierarchical recurrent neural
  networks.
\newblock In \emph{Proceedings of the IEEE Conference on Computer Vision and
  Pattern Recognition}.

\bibitem[{Yu et~al.(2017)Yu, Choi, Kim, Yoo, Lee, and Kim}]{HumanGaze}
Yu, Y.; Choi, J.; Kim, Y.; Yoo, K.; Lee, S.-H.; and Kim, G. 2017.
\newblock Supervising neural attention models for video captioning by human
  gaze data.
\newblock In \emph{Proceedings of the IEEE Conference on Computer Vision and
  Pattern Recognition}.

\bibitem[{Zhang and Peng(2019{\natexlab{a}})}]{OA-BTG}
Zhang, J.; and Peng, Y. 2019{\natexlab{a}}.
\newblock Object-aware Aggregation with Bidirectional Temporal Graph for Video
  Captioning.
\newblock In \emph{Proceedings of the IEEE Conference on Computer Vision and
  Pattern Recognition}.

\bibitem[{Zhang and Peng(2019{\natexlab{b}})}]{zhang2019object}
Zhang, J.; and Peng, Y. 2019{\natexlab{b}}.
\newblock Object-aware Aggregation with Bidirectional Temporal Graph for Video
  Captioning.
\newblock In \emph{Proceedings of the IEEE Conference on Computer Vision and
  Pattern Recognition}.

\bibitem[{Zhang et~al.(2017)Zhang, Gao, Zhang, Zhang, Li, and Tian}]{TDDF}
Zhang, X.; Gao, K.; Zhang, Y.; Zhang, D.; Li, J.; and Tian, Q. 2017.
\newblock Task-driven dynamic fusion: Reducing ambiguity in video description.
\newblock In \emph{Proceedings of the IEEE Conference on Computer Vision and
  Pattern Recognition}.

\bibitem[{Zhang et~al.(2020)Zhang, Shi, Yuan, Li, Wang, Hu, and Zha}]{ORG-TRL}
Zhang, Z.; Shi, Y.; Yuan, C.; Li, B.; Wang, P.; Hu, W.; and Zha, Z.-J. 2020.
\newblock Object Relational Graph with Teacher-Recommended Learning for Video
  Captioning.
\newblock In \emph{Proceedings of the IEEE/CVF Conference on Computer Vision
  and Pattern Recognition}.

\bibitem[{Zhao et~al.(2018)Zhao, Li, Lu et~al.}]{zhao2018video}
Zhao, B.; Li, X.; Lu, X.; et~al. 2018.
\newblock Video Captioning with Tube Features.
\newblock In \emph{Proceedings of the 27th International Joint Conference on
  Artificial Intelligence}.

\bibitem[{Zheng, Wang, and Tao(2020)}]{SAAT}
Zheng, Q.; Wang, C.; and Tao, D. 2020.
\newblock Syntax-Aware Action Targeting for Video Captioning.
\newblock In \emph{Proceedings of the IEEE/CVF Conference on Computer Vision
  and Pattern Recognition}.

\bibitem[{Zhou et~al.(2019)Zhou, Kalantidis, Chen, Corso, and Rohrbach}]{GVD}
Zhou, L.; Kalantidis, Y.; Chen, X.; Corso, J.~J.; and Rohrbach, M. 2019.
\newblock Grounded video description.
\newblock In \emph{Proceedings of the IEEE Conference on Computer Vision and
  Pattern Recognition}.

\end{thebibliography}

\end{document}